\documentclass{esannV2}
\usepackage{graphicx}
\usepackage[latin1]{inputenc}
\usepackage{amssymb,amsmath,array}
\usepackage{url}
\usepackage[table]{xcolor}
\usepackage{booktabs}
\usepackage{multicol,multirow}

\definecolor{Gray}{gray}{0.93}
\newcommand{\ptype}[1]{\mathcal{P}_{#1}}
\newcommand{\slgcn}{\mathcal{M}_{SL}}

\begin{document}
\title{Exploring Strategies for Modeling \\ Sign Language Phonology}

\author{Lee Kezar$^1$, Riley Carlin$^1$, Tejas Srinivasan$^1$, \\ Zed Sehyr$^2$, Naomi Caselli$^3$, and Jesse Thomason$^1$
%
%
\vspace{.3cm}\\
%
1 -- University of Southern California,
Los Angeles, California\\
%
2 -- Chapman University,
Orange, California\\
3 -- Boston University,
Boston, Massachusetts\\
}

\maketitle

\begin{abstract}
Like speech, signs are composed of discrete, recombinable features called phonemes. 
Prior work shows that models which can recognize phonemes are better at sign recognition, motivating deeper exploration into strategies for modeling sign language phonemes. 
In this work, we learn graph convolution networks to recognize the sixteen phoneme ``types'' found in ASL-LEX 2.0.
Specifically, we explore how learning strategies like multi-task and curriculum learning can leverage mutually useful information between phoneme types to facilitate better modeling of sign language phonemes. 
Results on the Sem-Lex Benchmark show that curriculum learning yields an average accuracy of 87\% across all phoneme types, outperforming fine-tuning and multi-task strategies for most phoneme types. 
\end{abstract}

\section{Introduction}

Phonology can act as a low-level yet discrete feature space to help guide a language model's perception of language.
This guidance is particularly attractive for computationally modeling signed languages, a task where accurate and reliable perception is fundamental but frequently muddied by insufficient data and a high degree of signer variation.
From the perspective of phonology, however, the features of interest are significantly easier to learn.
As the systematic components of signs, phonemes are by definition more abundant and less complex than whole signs.
Meanwhile, the utility of phoneme recognition for understanding signed language is clear.
\cite{kezar:islr_phonology} showed that leading models for isolated sign recognition (ISR) do not reliably encode sign language phonemes, but with supervision for phonemes alongside gloss, those models will be up to 9\% more accurate at ISR.
Moreover, the descriptive power of sign language phonology can readily extend to sign constructions not found in lexicons, like derivatives of signs (e.g. \textsc{day} vs. \textsc{two-days}) and classifier constructions (e.g. \textsc{CL:drive-up-hill}).

Building on these observations, we focus on modeling sign language phonology as a task unto itself.
We evaluate two learning strategies, multi-task and curriculum learning, on their ability to improve the recognition of American Sign Language (ASL) phonemes.
Our experiments using the Sem-Lex Benchmark~\cite{semassoc} to learn a graph convolution network reveal that learning phoneme types together (rather than separately) improves accuracy. We additionally show that curriculum learning, wherein the model is given structural priors related to phoneme types, is the most accurate method to date.

\section{Related Work on Modeling Sign Language Phonology}
Several related works have explored models for sign language phonology, both as its own task and in relation to sign recognition, in a variety of ways. Perhaps the earliest effort to recognize sign language phonemes, \cite{movhold1} explores the use of nearest-neighbor classifiers for recognizing handshapes, palm orientations, locations, and movements, based on hand-crafted feature representations of the hands and body, such as ``rotation values of the hand joints.''
Although they claim 85\%--95\% accuracy, the classifiers are trained and evaluated on synthetic sign recognition, raising concerns regarding their classifiers' ability to generalize to naturalistic signing.
\begin{table}[t!]
    \centering
    \begin{tabular}{llc} 
        \toprule
        \multicolumn{1}{c}{Phoneme Type}    & \multicolumn{1}{c}{Description} & \#Values \\
        \midrule
        Major Location                      & The sign's broad location. & 5 \\
        
        \rowcolor{Gray} Minor Location      & The signs's specific location. & 37 \\
        
        Second Minor Loc.                   & The sign's specific, secondary location. & 37 \\
        
        \rowcolor{Gray} Contact             & If the hand touches body. & 2 \\
        
        Thumb Contact                       & If the thumb touches other fingers. & 3 \\

        \rowcolor{Gray}Sign Type            & Movement symmetry (if 2H) & 6 \\

        Repeated Movement                   & If the movement is repeated. & 2 \\

        \rowcolor{Gray} Path Movement       & The shape that the hand traces. & 8  \\

        Wrist Twist                         & If the hand rotates. & 2 \\

        \rowcolor{Gray} Spread                              & If the hand's fingertips touch. & 3 \\

        Flexion             & The way the finger joints are bent. & 8  \\

        \rowcolor{Gray}Thumb Position       & If the thumb is in/out. & 2 \\

        Selected Fingers                    & Which fingers are salient to the sign. & 8 \\

        \rowcolor{Gray} Spread Change       & If \textit{Spread} changes. & 3 \\

        Nondom. Handshape                   & Configuration of the nondominant hand. & 56 \\
        
        \rowcolor{Gray} Handshape           & Configuration of the dominant hand. & 58 \\
        \bottomrule
    \end{tabular}
    \caption{Overview of each phoneme types found in ASL-LEX 2.0, including the number of possible values. See \cite{asllex} for a more detailed description of the types.}
    \label{intro:tab:ptype}
\end{table}

Later efforts to recognize SL phonemes would focus on designing neural architectures to replace the hand-crafted features with encodings. While \cite{camgoz}, \cite{deephand}, and \cite{bsl1k} improve sign recognition by more intentionally attending to the hands and mouth, one might describe their connection with language \textit{phonetic}, as they are more closely associated with continuous input-level features than they are with discrete and symbolic representations.
WLASL-LEX \cite{wlasl-lex} is conceptually similar to the work presented here. This work compared four classification models for each of the 6 phoneme types found in ASL-LEX 1.0, learned with WL-ASL dataset.
In contrast, the work presented here uses the Sem-Lex Benchmark \cite{semassoc}, which contains 10 additional phoneme types (see Table \ref{intro:tab:ptype} and approximately 300\% more sign videos to learn from.
Additionally, we explore learning strategies rather than model architectures.

\section{Methodology}
\begin{figure}
    \centering
    \includegraphics[width=0.8\linewidth]{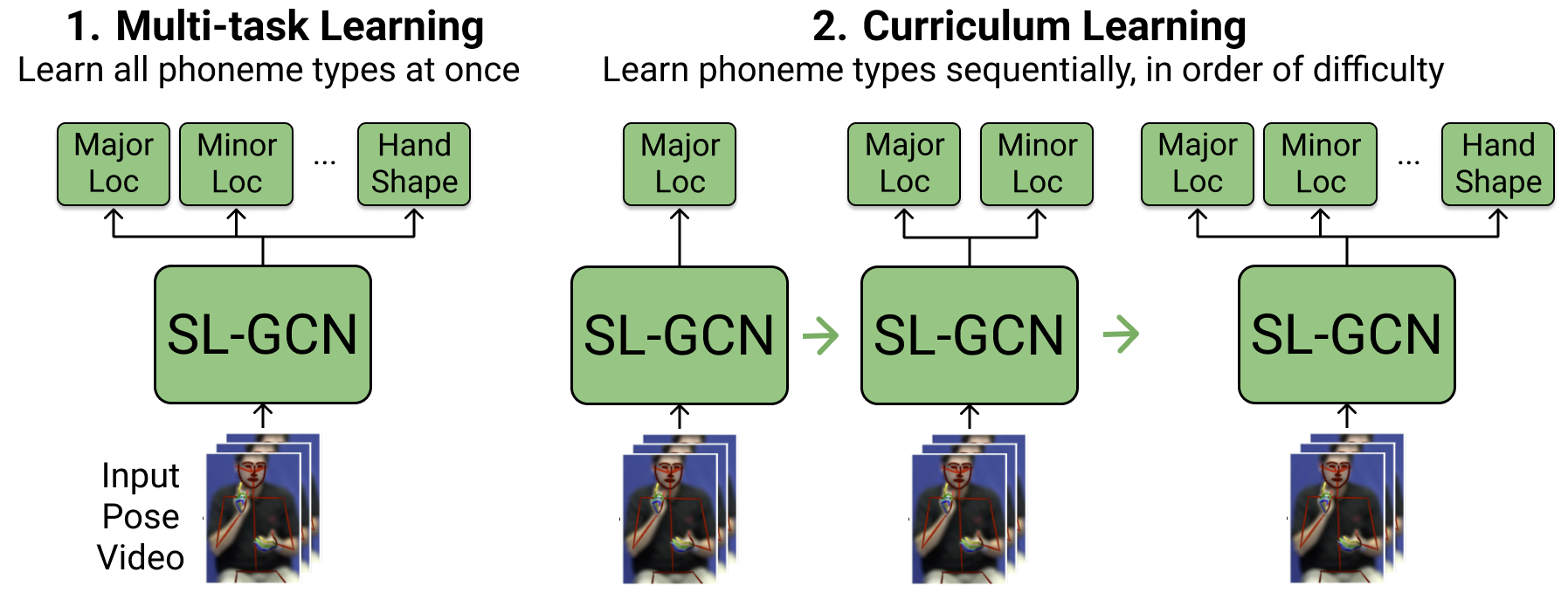}
    \caption{We explore multi-task and curriculum learning to improve modeling of sign language phonology by sharing knowledge across phoneme types.}
    \label{fig:methods}
\end{figure}

\subsection{Task Description}

Brentari's Prosodic Model~\cite{brentari} organizes sign language phonology into a hierarchy of sixteen distinct phoneme types $\ptype{1..16}$. 
We view learning each phoneme type $\ptype{i}$ as a classification task with $K_i$ distinct classes, where a model takes as input a pose estimation video $\mathbf{x}$ and predicts an output class $y \in \{1,...,K_i\}$.

\subsection{Learning to Classify Phoneme Types with SL-GCN}

Following~\cite{kezar:islr_phonology}, we perform phoneme classification using an SL-GCN encoder~\cite{SL-GCN} $\slgcn$ to encode the pose estimation video. 
To classify phoneme type $\ptype{i}$, a linear classification layer $\theta_i$ maps the encoding to a probability distribution $p(y | \mathbf{x}; \slgcn, \theta_i)$ over the $K_i$ output classes of that phoneme type. 
The cross-entropy loss with ground-truth label $\mathbf{y}_i$ is minimized over training dataset $\mathcal{D}$:
\begin{align}
    \min_{\mathbf{x}, \mathbf{y}_i \sim \mathcal{D} } \mathcal{L}_{CE} \Bigl( \mathbf{y}_i, \hspace{0.5em} p(y | \mathbf{x}; \slgcn, \theta_i) \Bigr)
\end{align}

\subsection{Multi-task Learning of Phoneme Types}

Training separate models for each phoneme type misses an opportunity to leverage shared knowledge across phoneme types. 
To this end, the first strategy we explore is multi-task learning of phoneme types, where individual classification layers for each of the 16 phoneme types are trained simultaneously.
All 16 phoneme type classifiers $\theta_{1...16}$ are learned jointly using video encodings from a shared SL-GCN encoder.
\begin{align}
    \min_{\mathbf{x}, \mathbf{y}_{1...16} \sim \mathcal{D} } \sum_{i=1}^{16} \mathcal{L}_{CE} \Bigl( \mathbf{y}_i, \hspace{0.5em} p(y | \mathbf{x}; \slgcn, \theta_i) \Bigr)
\end{align}

\newpage
\subsection{Curriculum Learning of Phoneme Types}

While multi-task learning allows the model to implicitly share knowledge across phoneme types, there is no structural prior or inductive bias that regulates how the knowledge is shared. 
Controlling the order in which phoneme types are introduced might introduce such a structural prior.
For instance, learning to locate the hands first can help us identify the type of hand movement better. 

To decide this order, we follow two principles: earlier types should be ``easier'' than later types, and the knowledge of earlier types should reduce the entropy of later types. Because Brentari's Prosodic Model is hierarchical---phoneme types have children and/or parent types---the most sensible way to follow these principles is to start with ``leaf'' phoneme types (those which have no children and fewer values) and moving up towards broader, more holistic phoneme types. For example, Handshape has children types Flexion, Selected Fingers, et al. Ergo, learning the more specific children types before Handshape is both easier (in terms of number of values possible values) and reduces the entropy of Handshape. The resulting curriculum is shown in the ordering of Table \ref{intro:tab:ptype}, starting with Major Location and ending in Handshape.

We perform curriculum learning by introducing phoneme types into the learning objective cumulatively.
We begin training by only learning phoneme type $\ptype{1}$, and introduce a new phoneme type $\ptype{k}$ into the learning objective every $e$ epochs. 
For the final $e$ epochs, model training is identical to multi-task learning of all 16 phoneme types $\ptype{1...16}$.
\begin{align}
    \text{Step $k$}: \min_{\mathbf{x}, \mathbf{y}_{1...k} \sim \mathcal{D} } \sum_{i=1}^{k} \mathcal{L}_{CE} \Bigl( \mathbf{y}_i, \hspace{0.5em} p(y | \mathbf{x}; \slgcn, \theta_i) \Bigr)
\end{align}


\section{Data and Experimental Setup}
To evaluate our method, we use the Sem-Lex Benchmark~\cite{semassoc}, which contains 65,935 isolated sign videos annotated by humans with both gloss and ASL-LEX phoneme types.
This dataset was collected from deaf, fluent signers who gave informed consent and received financial compensation.
We use the train partition ($n=51,029$) gloss labels to pre-train the SL-GCN model to recognize gloss only and use this as the base model to fine-tune for phonological feature recognition.
For multi-task learning, we use a cosine-annealing learning rate and train for 100 epochs, at which point the validation accuracy plateaus.
For curriculum learning, we follow the same procedure but with $e=20$ between the introduction of a new phoneme type. 
Models are implemented in PyTorch, largely building on the OpenHands framework \cite{openhands}, and trained on four Nvidia 3090 GPUs. Our code can be found at \url{https://github.com/leekezar/Modeling-ASL-Phonology/}.

\newpage 
\section{Results and Discussion}
\begin{table}[t!]
    \centering
    \begin{tabular}{lcccc} 
        \toprule
        \multicolumn{1}{c}{\multirow{2}{*}{\textbf{Phoneme Type}}} & \multicolumn{3}{c}{\textbf{Learning Method}} & \multirow[c]{2}[2]{*}{\begin{tabular}{@{}c@{}}Type \\ Average\end{tabular}} \\
        \cmidrule{2-4}
        \multirow{2}{*}{} & Fine-Tune & Multitask & Curriculum & \\
        \midrule
        Major Location                          & 87.7 & 87.5	& \textbf{89.1} & 88.1 \\
        \rowcolor{Gray} Minor Location          & 79.2 & 78.1	& \textbf{80.7} & 79.3 \\
        Second Minor Location                   & 78.7 & 77.2	& \textbf{80.9} & 78.9\\
        \rowcolor{Gray} Contact                 & 89.3 & 88.6	& \textbf{91.1} & 89.7\\
        Thumb Contact                           & 91.7 & 91.1	& \textbf{92.1} & 91.6\\
        \rowcolor{Gray} Sign Type               & 88.9 & 87.9	& \textbf{89.4} & 88.7\\
        Repeated Movement                       & 85.5 & 85.4	& \textbf{87.3} & 86.1\\
        \rowcolor{Gray} Path Movement           & 75.6 & 75.4	& \textbf{79.6} & 76.9\\
        Wrist Twist                             & 92.4 & 92.6	& \textbf{93.5} & 92.8\\
        \rowcolor{Gray} Selected Fingers        & \textbf{91.1} & 90.2 & 90.6 & 90.6\\
        Thumb Position                          & 91.5 & 91.5	& \textbf{91.8} & 91.6\\
        \rowcolor{Gray} Flexion                 & 81.2 & 81.0	& \textbf{83.2} & 81.8\\
        Spread                                  & 88.4 & 88.0	& \textbf{88.8} & 88.4\\
        \rowcolor{Gray} Spread Change           & 90.3 & 89.5	& \textbf{90.4} & 90.1\\
        Nondominant Handshape                   & \textbf{83.5} & 81.7 & 83.2 & 82.8\\
        \rowcolor{Gray} Handshape               & \textbf{77.4} & 74.7 & 76.9 & 76.3\\
        \midrule
        Method Average                                 & 85.8 & 85.0 & \textbf{86.8} & 85.9\\
        \bottomrule
    \end{tabular}
    \caption{Phoneme recognition top-1 accuracy (\%) across the proposed methods, evaluated on Sem-Lex (test). All models are pre-trained to predict sign gloss.}
    \label{results:tab:phon}
\end{table} \label{results}

The top-1 accuracies for each phoneme type across methods are shown in Table \ref{results:tab:phon}.
Overall, the three methods are effective at learning the phonological features in Sem-Lex, with an overall accuracy of $85.9\%$.
This outperforms WLASL-LEX \cite{wlasl-lex} across its six phoneme types by $5.9$--$20.9\%$.
From these results, we glean the following conclusions:
\begin{itemize}
    \item \textbf{Phoneme types co-occur.} There is a relatively small difference of 0.8\% between learning the entire model for each phoneme type individually (fine-tune) vs. learning them all at once (multi-task). This indicates that the value of $\ptype{i}$ informs the value of $\ptype{j}$ to such an extent that it overcomes the challenges associated with learning many tasks simultaneously.

    \item \textbf{Inductive priors help.} The slight but consistent improvement imbued by the curriculum shows that, in addition to co-occurrence (captured by the multi-task strategy), there exist structural priors in the form of hierarchical relationships. In other words, the information gain is minimized (i.e. $\ptype{i}$ is least surprising) when more fine-grained phoneme types are learned \textit{after} coarse-grained ones.
\end{itemize}

\section{Conclusion}
In this work, we provide empirical evidence that modeling sign language phonology is a complex task which benefits from special attention to linguistic theory.
By learning models from high-quality, specialized data which reflect phonological features in sign language, we show that phonemes exhibit both co-occurrence and hierarchical relationships. Future work will compare varied curricula, explore the capacity of phonemes to describe a variety of sign constructions, and assess any biases associated with race and gender.



\begin{footnotesize}



\bibliographystyle{unsrt}
\bibliography{bibliography}

\begin{thebibliography}{10}

\bibitem{kezar:islr_phonology}
Lee Kezar, Jesse Thomason, and Zed~Sevcikova Sehyr.
\newblock Improving sign recognition with phonology.
\newblock In {\em European Chapter of the ACL (EACL)}, 2023.

\bibitem{semassoc}
Lee Kezar, Elana Pontecorvo, Adele Daniels, Connor Baer, Ruth Ferster, Lauren
  Berger, Jesse Thomason, Zed Sehyr, and Naomi Caselli.
\newblock {The Sem-Lex Benchmark: Modeling ASL Signs and Their Phonemes}.
\newblock In {\em ACM SIGACCESS Conference on Computers and Accessibility
  (ASSETS)}, 2023.

\bibitem{movhold1}
Kabil Jaballah.
\newblock Towards content-based 3d sign language indexing using segmental m-h
  model.
\newblock {\em Fourth International Conference on Information and Communication
  Technology and Accessibility (ICTA)}, pages 1--4, 2013.

\bibitem{asllex}
Zed~Sevcikova Sehyr, Naomi~K. Caselli, Ariel Cohen-Goldberg, and Karen Emmorey.
\newblock {The ASL-LEX 2.0 Project: A Database of Lexical and Phonological
  Properties for 2,723 Signs in American Sign Language}.
\newblock {\em The Journal of Deaf Studies and Deaf Education}, 2021.

\bibitem{camgoz}
Necati Cihan~Camgoz, Simon Hadfield, Oscar Koller, and Richard Bowden.
\newblock Subunets: End-to-end hand shape and continuous sign language
  recognition.
\newblock In {\em International Conference on Computer Vision (ICCV)}, 2017.

\bibitem{deephand}
Ayan Sinha, Chiho Choi, and Karthik Ramani.
\newblock Deephand: Robust hand pose estimation by completing a matrix imputed
  with deep features.
\newblock In {\em Computer Vision and Pattern Recognition (CVPR)}, 2016.

\bibitem{bsl1k}
Samuel Albanie, G{\"u}l Varol, Liliane Momeni, Triantafyllos Afouras, Joon~Son
  Chung, Neil Fox, and Andrew Zisserman.
\newblock {BSL-1K}: {S}caling up co-articulated sign language recognition using
  mouthing cues.
\newblock In {\em European Conference on Computer Vision}, 2020.

\bibitem{wlasl-lex}
Federico Tavella, Viktor Schlegel, Marta Romeo, Aphrodite Galata, and Angelo
  Cangelosi.
\newblock {WLASL}-{LEX}: a dataset for recognising phonological properties in
  {A}merican {S}ign {L}anguage.
\newblock In {\em ACL}, 2022.

\bibitem{brentari}
Diane Brentari.
\newblock {\em A Prosodic Model of Sign Language Phonology}.
\newblock The MIT Press, 1998.

\bibitem{SL-GCN}
Songyao Jiang, Bin Sun, Lichen Wang, Yue Bai, Kunpeng Li, and Yun~Raymond Fu.
\newblock Skeleton aware multi-modal sign language recognition.
\newblock {\em 2021 IEEE/CVF Conference on Computer Vision and Pattern
  Recognition Workshops (CVPRW)}, 2021.

\bibitem{openhands}
Prem Selvaraj, C.~GokulN., Pratyush Kumar, and Mitesh~M. Khapra.
\newblock {OpenHands: Making Sign Language Recognition Accessible with
  Pose-based Pretrained Models across Languages}.
\newblock In {\em ACL}, 2022.

\end{thebibliography}

\end{footnotesize}


\end{document}